\documentclass{article}

\usepackage{arxiv}

\usepackage[utf8]{inputenc} %
\usepackage[T1]{fontenc}    %
\usepackage{hyperref}       %
\usepackage{url}            %
\usepackage{booktabs}       %
\usepackage{amsfonts}       %
\usepackage{nicefrac}       %
\usepackage{microtype}      %
\usepackage{lipsum}		%
\usepackage{graphicx}
\usepackage{doi}
\usepackage{units}

\title{Foundation Models for Structural Health Monitoring}

\author{Luca Benfenati\\
	DAUIN, Politecnico di Torino\\
    Turin, 10125 \\
	\texttt{luca.benfenati@polito.it} \\
	\And
	Daniele Jahier Pagliari\\
	DAUIN, Politecnico di Torino\\
    Turin, 10125 \\
	\texttt{daniele.jahier@polito.it} \\
    \And
    Luca Zanatta\\
    DEI, University of Bologna\\
    Bologna, 40136\\
    \texttt{luca.zanatta3@unibo.it}
    \And
    Yhorman Alexander Bedoya Velez\\
	DAUIN, Politecnico di Torino\\
    Turin, 10125 \\
	\texttt{s287296@studenti.polito.it} \\
    \And
    Andrea Acquaviva\\
    DEI, University of Bologna\\
    Bologna, 40136\\
    \texttt{andrea.acquaviva@unibo.it}\\
    \And
    Massimo Poncino\\
	DAUIN, Politecnico di Torino\\
    Turin, 10125 \\
	\texttt{massimo.poncino@polito.it} \\
    \And
    Enrico Macii\\
    DIST, Politecnico di Torino\\
    Turin, 10125 \\
	\texttt{enrico.macii@polito.it} \\
    \And
    Luca Benini\\
    D-ITET, ETH Zurich\\
    Zürich, 8092\\
    \texttt{lbenini@iis.ee.ethz.ch}\\
    \And
    Alessio Burrello\\
    DIST, Politecnico di Torino\\
    Turin, 10125 \\
	\texttt{alessio.burrello@polito.it} \\   
}

\date{}

\renewcommand{\shorttitle}{\textit{arXiv} Template}

\hypersetup{
pdftitle={Foundations Models on Structural Health Monitoring},
pdfsubject={cs.Artificial-Intelligence},
pdfauthor={Benfenati Luca},
pdfkeywords={Structural Health Monitoring, Traffic Load Estimation, Foundation Models, Masked Autoencoders, Deep Learning},
}

\begin{document}

\maketitle

\begin{abstract}
Structural Health Monitoring (SHM) is a critical task for ensuring the safety and reliability of civil infrastructures, typically realized on bridges and viaducts by means of vibration monitoring.
In this paper, we propose for the first time the use of Transformer neural networks, with a Masked Auto-Encoder architecture, as \textit{Foundation Models} for SHM. We demonstrate the ability of these models to learn generalizable representations from multiple large datasets through self-supervised pre-training, which, coupled with task-specific fine-tuning, allows them to outperform state-of-the-art traditional methods on diverse tasks, including Anomaly Detection (AD) and Traffic Load Estimation (TLE). We then extensively explore model size versus accuracy trade-offs and experiment with Knowledge Distillation (KD) to improve the performance of smaller Transformers, enabling their embedding directly into the SHM edge nodes.

We showcase the effectiveness of our foundation models using data from three operational viaducts. For AD, we achieve a near-perfect 99.9\% accuracy with a monitoring time span of just 15 windows. In contrast, a state-of-the-art method based on Principal Component Analysis (PCA) obtains its first good result (95.03\% accuracy), only considering 120 windows. On two different TLE tasks, our models obtain state-of-the-art performance on multiple evaluation metrics (R$^2$ score, MAE\% and MSE\%). On the first benchmark, we achieve an R$^2$ score of 0.97 and 0.90 for light and heavy vehicle traffic, respectively, while the best previous approach (a Random Forest) stops at 0.91 and 0.84. On the second one, we achieve an R$^2$ score of 0.54 versus the 0.51 of the best competitor method, a Long-Short Term Memory network.
\end{abstract}


\section{Introduction}\label{sec:introduction}

Structural Health Monitoring (SHM) is crucial to ensure longevity and safety of critical infrastructures such as bridges, highways, and tunnels. In fact, these structures are subject to deterioration due to factors such as ageing, environmental conditions, and traffic loads, posing potential safety hazards and substantial financial burdens for maintenance and repairs.

SHM systems address this challenge by proactively and continuously monitoring infrastructure in real time, enabling early detection of anomalies and potential damages. This facilitates timely interventions that ensure public safety while optimizing resource allocation through cost-effective strategies.
When applied to critical infrastructures such as bridges and viaducts, SHM can also exploit Traffic Load Estimation (TLE) as a metric that captures the dynamic nature of traffic loads that the structures endure, to continuously assess structural health and predict maintenance needs.

A typical SHM system comprises a network of sensors responsible for measuring multiple parameters relevant to the current state of the structure as well as its surrounding environment. The most widely used SHM sensing devices include fiber optic sensors~\cite{fibersensor, odat2017vehicle}, accelerometers~\cite{accelerometer, accelerometer2}, cameras~\cite{kamkar2016vehicle} and strain gauges~\cite{straingauges}.
Among these options, networks of Micro-Electro-Mechanical Systems (MEMS) for vibration monitoring stand out due to their cost-effectiveness and relatively simple deployment in diverse environments.
Data gathered by these accelerometers are then used as the input of a Machine Learning (ML) model to address tasks such as Anomaly Detection (AD) or TLE. These devices have demonstrated high accuracy compared to other technologies on several SHM tasks~\cite{sensors_shm_tasks}.

ML approaches for vibration-based SHM span a diverse range of techniques, from classic methods as in\cite{burrello2022traffic, Finotti2019AnSA}, to more recent deep learning solutions such as Convolutional Neural Networks (CNNs) \cite{cnn_ad} and Autoencoders \cite{autoencoder1, autoencoder2}. While Transformer neural networks have revolutionized the Computer Vision and Natural Language Processing (NLP) fields, in SHM they have only been applied to camera-based monitoring solutions \cite{transformer_shm}, and not yet (to our knowledge) to vibration-based ones. 

In this paper, we introduce two key contributions in the direction of a new family of models for acceleration-based SHM. First, we explore, for the first time, the application of Transformers to this task. Second, building on the recent breakthroughs of foundation models across diverse domains (language, vision, etc.)~\cite{foundation_model}, we explore self-supervised learning to take advantage of large amounts of easily available unlabelled data and develop general backbone feature extractors, which can be then fine-tuned for multiple SHM tasks. By doing so, we outperform current, single-task ML approaches, which are instead trained on small datasets in a fully supervised way.

More specifically, we adopt a transformer-based Masked Autoencoder architecture inspired by \cite{huang2022masked}, which has become a standard in self-supervised learning, demonstrating better performance compared to classical autoencoder architectures.
This approach demonstrates the potential of foundation models for advancing data-driven SHM, particularly in overcoming the challenge of collecting large amounts of labeled data, a process that is typically time-consuming and costly, requiring extensive instrumentation of target infrastructures with additional sensors, such as cameras or Weigh-in-Motion (WiM) systems \cite{wim} to obtain ground truth data.

In summary, the contributions of this work are:
\begin{itemize}
    \item We consider three different SHM datasets, including a newly collected one, larger than those introduced in previous works and fully labelled. These datasets are collected by a specialized company with patented sensor networks, expert in real-time SHM. They include two viaducts and a bridge, representative of common Italian infrastructures.
    \item We build a Transformer-based Masked Autoencoder, inspired by \cite{huang2022masked}, as the first Foundation Model for SHM, capable of addressing multiple tasks within a unified framework. By pre-training on three diverse datasets without labels (self-supervised learning) and fine-tuning on each task, we outperform individually trained models, demonstrating the benefits of transferable representations. This two-stage training approach simulates the realistic SHM scenario where large unlabelled datasets are easily available, but labelled samples are scarce.
    \item Through our experiments, our fine-tuned models outperform state-of-the-art algorithms on all three datasets. In particular, on the first dataset, we achieve an AD accuracy of 99.92\%, a sensitivity of 100\%, and a specificity of 99.9\%, with respect to state-of-the-art 75.76\%, 55.68\%, and 98.75\%. On the other two datasets, and on three TLE variants, we achieve an R$^2$ score of 0.97, 0.85, and 0.54 respectively, outperforming the state-of-the-art that stops at 0.91, 0.84, and 0.10.
    \item We carry out an extensive search on the optimal model size. In this context, we test Knowledge Distillation (KD) to train smaller models to imitate larger ones, ultimately targeting deployment on resource-constrained nodes for real-time SHM at the edge. Results show that distilled models often outperform standardly fine-tuned and equally sized counterparts on downstream tasks.
    \item We demonstrate the feasibility of deploying our models on a gateway-class edge device, i.e., the NVIDIA Jetson Nano, showing a latency well below the real-time constraint of the three analyzed use cases.
\end{itemize}
The rest of the paper is organized as follows. Section \ref{sec:background} provides the necessary background. Section \ref{sec:related} overviews the SHM literature, describing the different approaches to AD and TLE. Section \ref{sec:installation} describes the three considered viaducts, the corresponding SHM sensor networks, and the data acquisition and labeling processes. Section \ref{sec:methodology} details our foundation model approach, presenting the processing pipeline, the architecture of the model, and its training procedure. Section \ref{sec:results} presents the experimental results and Section \ref{sec:conclusions} concludes the paper.
We open-source our code and pre-trained models at \texttt{\url{https://github.com/eml-eda/tle-supervised}}.
\section{Background}\label{sec:background}
\subsection{Transformer Neural Networks}
\begin{figure}
    \centering
    \includegraphics[width=0.5\textwidth]{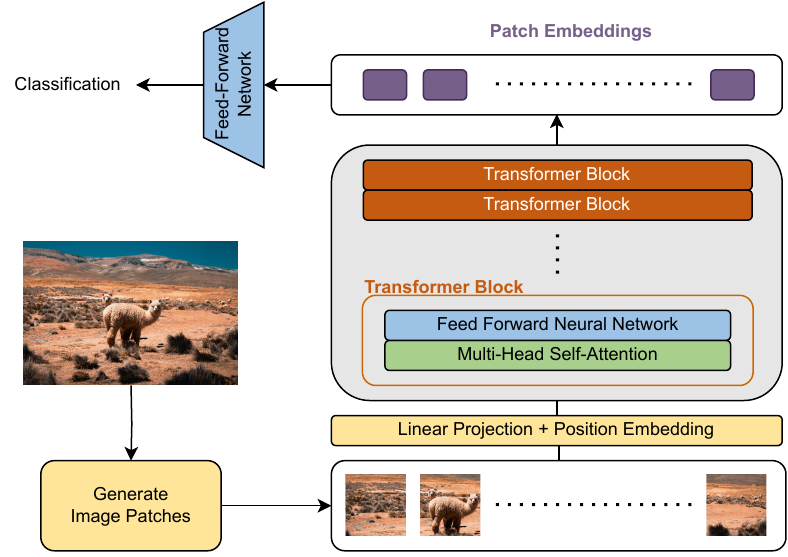}
    \caption{Vision Transformer ViT Architecture}
    \label{fig:vit}
\end{figure}

The Transformer architecture was introduced in 2017~\cite{transformer}, ushering in a new era for NLP applications and DL in general. The architecture leverages an encoder-decoder structure with stacked self-attention and feedforward layers, enabling the model to process input sequences and generate outputs (e.g., translations) autoregressively. Its core innovation lies in self-attention, which allows the model to directly attend to relevant parts of the input sequence, capturing long-range dependencies. Scaled dot-product and Multi-Head Self-Attention (MHSA) enable efficient and diverse attention computation, while positional encodings address the lack of inherent sequence awareness~\cite{mhsa}. Its impact goes beyond NLP, making it a cornerstone of modern DL due to its scalability, parallelizability, and ability to model long-range dependencies. In our work, we use an architecture derived from the Vision Transformer (ViT) ~\cite{vit_transformers}, which first applied an attention-based Transformer to image processing. 
This is achieved by breaking the image into fixed-size patches, linearly embedding each of them, adding the position embeddings and then feeding the resulting sequence of vectors to a standard Transformer encoder. A simplified scheme of a ViT encoder architecture is shown in Figure \ref{fig:vit}. Notably, we apply a similar architecture to vibration data rather than standard RGB images, after extracting their spectrogram, as detailed in Sec.~\ref{sec:pipeline}.

\subsection{Masked Autoencoders}
Masked autoencoders sparked renewed interest in 2021 when \cite{mae} successfully adopted them as scalable self-supervised learners for Computer Vision applications. They are encoder-decoder models, whose encoder is the ViT shown in the previous section. The decoder, instead, takes the patch embeddings as inputs and is tasked to reconstruct the original image. During training, some image patches are randomly masked before being fed to the network, forcing the autoencoder to reconstruct the missing information solely from the remaining visible parts. This process encourages the model to learn robust and diverse representations of visual features. The simplicity and effectiveness of this scheme have led masked autoencoders to achieve state-of-the-art performance in various downstream tasks, demonstrating their potential as powerful tools for visual representation learning~\cite{mhsa}. The specific masked autoencoder architecture used in our work will be detailed in Sec.~\ref{sec:model} and shown in Fig.~\ref{fig:base_model}.

\subsection{Foundation Models}
The term Foundation Model refers to a DL model that is trained on broad data, using self-supervision at scale, and can be adapted to a wide range of downstream tasks~\cite{foundation_model}. The pre-training phase is usually carried out on diverse and unlabelled data, to produce general-purpose data representations that, unlike specialized models trained for specific tasks, serve as a foundational block. Fine-tuned on specific downstream tasks, foundation models offer several advantages: increased training efficiency, broader applicability, and the ability to transfer learned knowledge across domains. This shift in AI from task-specific models to a reusable ``foundation'' approach has opened new avenues for leveraging the power of DL across various applications, from tasks like summarization and translation in NLP\cite{foundation_nlp} to image classification and object detection in Computer Vision\cite{foundation_cv1, foundation_cv2}, and time-series forecasting \cite{fomo_timeseries_survey, moment}. However, the adoption of this novel paradigm in other domains, including SHM, is still unexplored.

\section{Related Works}\label{sec:related}
This section presents the current state-of-the-art approaches to SHM, including both AD (Sec.~\ref{subsec:related_1}) and TLE (Sec.~\ref{subsec:related_2}). For each task, we first broadly overview works that use other sensors, and then focus on accelerometer-based methods.

\subsection{Anomaly Detection}\label{subsec:related_1}
Concerning visual AD, \cite{transformer_shm} applies a ViT-like approach for semantic segmentation to identify damaged structural components in a railway viaduct, achieving 97\% and 90\% mIoU for component and damage segmentation, respectively, and outperforming existing vision-based methods. Similarly, \cite{unsupervised_ad_laboratory} detects bridge damage using a roving multi-camera system, comparing five unsupervised techniques (Autoencoder, K-Nearest Neighbours, Kernel Density, Local Outlier Factor, and Isolation Forest) and achieving F1 scores between 0.96 and 0.97 in controlled tests. While visual approaches reach impressive accuracy, they incur high installation costs and are effective only under good visibility conditions (e.g., not during night or foggy days).

Therefore, most AD methods for SHM rely on acceleration data. For example, \cite{1dcnn_accelerometer} employs a 1-dimensional CNN to estimate the Probability of Damage across nine scenarios of increasing severity. In \cite{cnn_ad}, statistical features (maximum, minimum, mean, variance, skewness, kurtosis) extracted from acceleration data feed into a CNN with three convolutional layers, global average pooling, and softmax, achieving 97.6\% accuracy on a long-span cable-stayed bridge dataset. Spiking neural networks have also been explored in \cite{zanatta2021damage, barchi2021spiking}, reaching 95\% accuracy and Matthews Correlation Coefficient (MCC) values of 0.88 for viaduct damage classification.
These supervised approaches require extensive labeled data, limiting their scalability. Furthermore, they are specifically trained on a single infrastructure, while their generalization over multiple ones is not tested. 

In contrast, unsupervised methods such as \cite{ae_anomalydetection_related} use Autoencoders to extract features from vibration responses for indirect damage diagnosis, though predictions can deviate by an average of five levels on a 30-level scale. Similarly, \cite{eltouny2023large} employs an unsupervised Recurrent Neural Network (RNN) for damage detection and localization, achieving 93\% and 85\% accuracy, respectively, albeit on simulated data.
The most promising unsupervised AD method for vibration data is presented in \cite{9729869}, which leverages PCA compression without labeled anomalies. PCA coefficients are fitted solely on healthy data, so anomalous vibrations are poorly reconstructed and flagged when the reconstruction error exceeds a threshold. On one of our real-world datasets—collected from a viaduct in northern Italy—this approach achieved 98.8\% accuracy, 100\% specificity, and 97.33\% sensitivity, and is used as the state-of-the-art benchmark in our comparisons (see Section \ref{sec:results}).

\subsection{Traffic Load Estimation}\label{subsec:related_2}
Kamkar et al.~\cite{kamkar2016vehicle} employ smart cameras to detect, classify, and quantify vehicles on public bridges, achieving 92.1\% accuracy in high-traffic scenarios. However, their performance falls to 74.8\% under low-light conditions, highlighting the sensitivity of visual-based methods to environmental factors. In contrast, methods using magnetic sensors with adaptive thresholds and classification trees~\cite{dong2018improved, wang2017roadside} report accuracy ranging from 84.7\% to 99.9\%, but they often achieve their best performance under controlled conditions (e.g., single, slow-moving vehicles) that do not reflect real-world traffic. Alternative sensor technologies, such as fiber optic~\cite{fibersensor} and infrared sensors~\cite{odat2017vehicle}, can also attain high accuracy; however, they also face challenges related to environmental sensitivity and high installation costs (they have to be embedded in the asphalt, requiring road interventions), limiting their feasibility in less populated areas. \cite{traffic_signal} develops an agent for adjusting traffic signals based on TLE data, but this approach does not directly address vehicle classification or quantification challenges.
On the other hand, acceleration-based approaches for TLE are less common. In \cite{tle_accelerometer}, the superiority of MEMS accelerometers is demonstrated, showing robustness to sensor placement and structural damage. In \cite{9128641}, TLE is framed as an anomaly detection problem, using k-means clustering to differentiate between light and heavy vehicles based on deviations from a periodically updated baseline. Although this method avoids reliance on labeled data, its performance in high-traffic conditions is insufficient. The most recent work, \cite{burrello2022traffic}, compares several classical machine learning techniques and a deep learning approach (Multi-Layer Perceptron) on MEMS accelerometer data, achieving state-of-the-art R$^2$ scores of 0.91 for light vehicles and 0.84 for heavy vehicles. We use this as the benchmark for comparison in Section~\ref{sec:results}.
\section{Structural Health Monitoring Installations}
\label{sec:installation}
This section describes the three SHM Use Cases (UCs) considered in our work. The first UC targets anomaly detection, identifying abrupt structural changes, while the other two focus on traffic load estimation with continuous targets. Whereas these three scenarios only regard vehicle communication infrastructures, they serve as a preliminary validation of self-supervised transformer-based models for large-scale SHM systems, which could be extended in the future to other civil structures whose structural health is measured by similar acceleration signals.

\subsection{Use Case 1 (UC1): Anomaly Detection Task}\label{subsec:uc1}
This UC utilizes vibration data from a dual-carriageway state highway viaduct in northern Italy. In 2019, a structural strengthening intervention on one of its spans altered the vibration signals. Before the intervention, five SHM nodes were installed under the bridge to monitor its vibrations, providing data from both pre- and post-intervention periods.

We treat post-intervention vibrations as normal and pre-intervention vibrations as anomalous, simulating abrupt structural changes (e.g., due to an earthquake). The sensor nodes use LIS344ALH 3-axis accelerometers \cite{lis344alh} (±2/±6g range) and an HTS221 temperature/humidity sensor \cite{hts221}. Data collection and processing are handled by an STM32L476VGTx microcontroller (ARM Cortex-M4, 80 MHz, 96 KB SRAM, 1 MB Flash).

\subsection{Traffic Load Estimation Task}
\subsubsection{Use Case 2 (UC2)}
\begin{figure}
    \centering
    \includegraphics[width=0.5\textwidth]{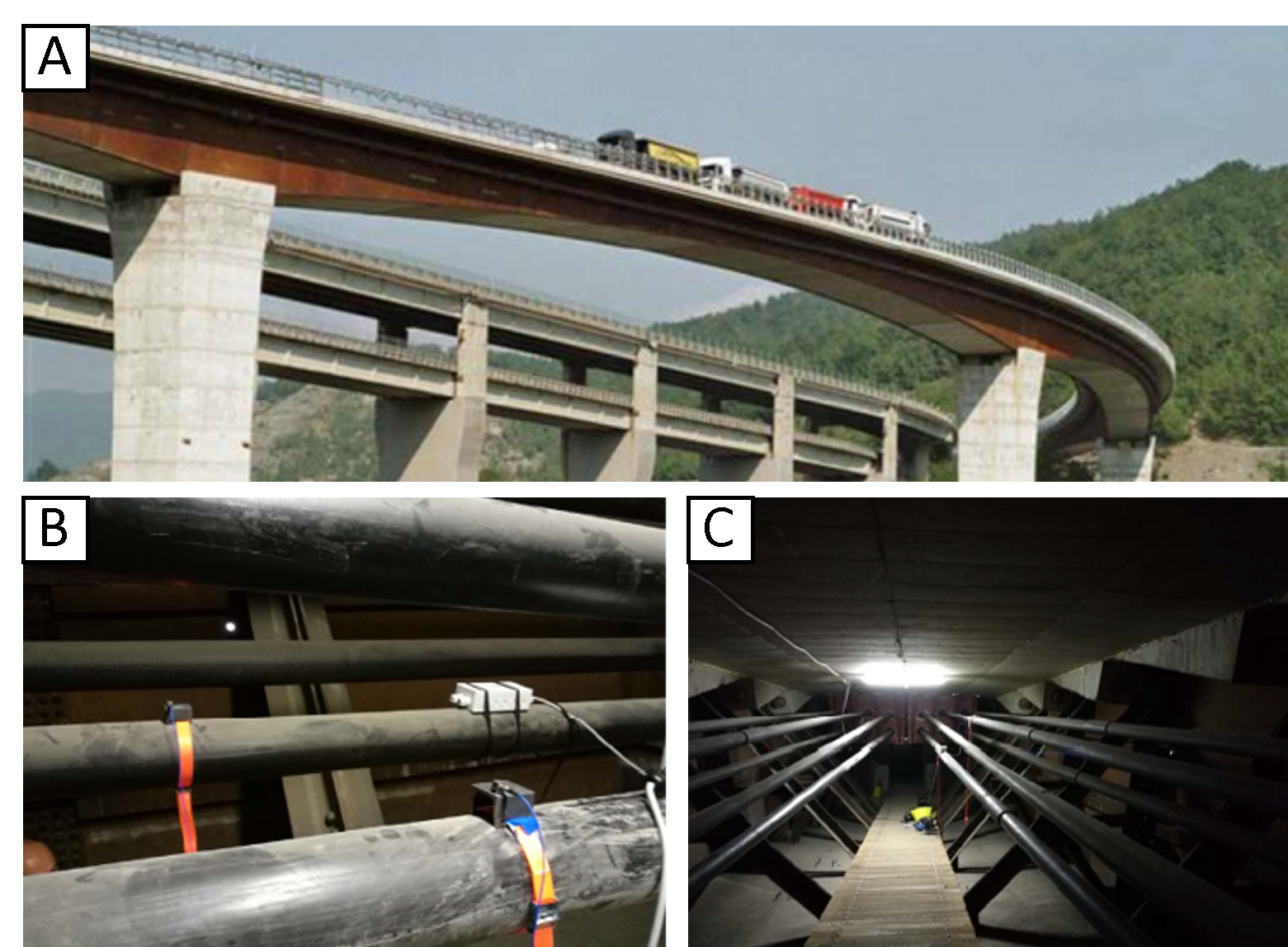}
    \caption{A) Aerial view of the viaduct under analysis. B) Sensor positioning on prestressing cables. C) Prestressing cables. }
    \label{fig:Roccaprebalza_viaduct}
\end{figure}
This UC focuses on a viaduct in Italy, monitored by an SHM sensor network since 2017 (Fig. \ref{fig:Roccaprebalza_viaduct}). The viaduct spans ~580 m, supported by five concrete pillars, with individual spans ranging from 67 m to 112 m. Each span consists of a steel caisson housing 12 prestressing cables, with cross-section heights varying from 6.0 m at the bearings to 3.0 m at mid-span.  

Given that the structural integrity of prestressed bridges depends on cable durability, a sensor system (Fig. \ref{fig:Roccaprebalza_viaduct}) was installed to detect breakages. Since September 2017, a network of 90 sensors has been actively collecting and storing data in the cloud. Each sensor node features an STM32F405RG microcontroller (MCU) for data acquisition and optional analytics, alongside the same accelerometer, temperature, and humidity sensors used in Section \ref{subsec:uc1}.

\subsubsection{Use Case 3 (UC3)}
This UC examines a reinforced concrete girder bridge with 18 spans, totaling 583 meters in length, with two lanes. The bridge follows an isostatic static scheme, with 20-meter spans, except for the first (10 m) and last (29.5 m) spans. Since 2022, an industrial SHM system has continuously recorded data.

The system includes 282 MEMS biaxial inclinometers and 142 MEMS triaxial accelerometers (±2 g full scale, 100 Hz sample rate), evenly distributed across spans. Accelerometers monitor the two external beams per span, with sensors positioned at the quarter, third, and midspan locations.

Additionally, a Weigh-in-Motion (WiM) system is installed 600 meters before the bridge, providing ground truth traffic data—including lane, vehicle length, weight, speed, and axle count. To align WiM data with SHM sensor readings, a time offset correction is applied based on vehicle speed and distance between the WiM system and each sensor.
\section{Methodology}\label{sec:methodology}
\begin{figure*}
    \centering
    \includegraphics[width=0.99\textwidth]{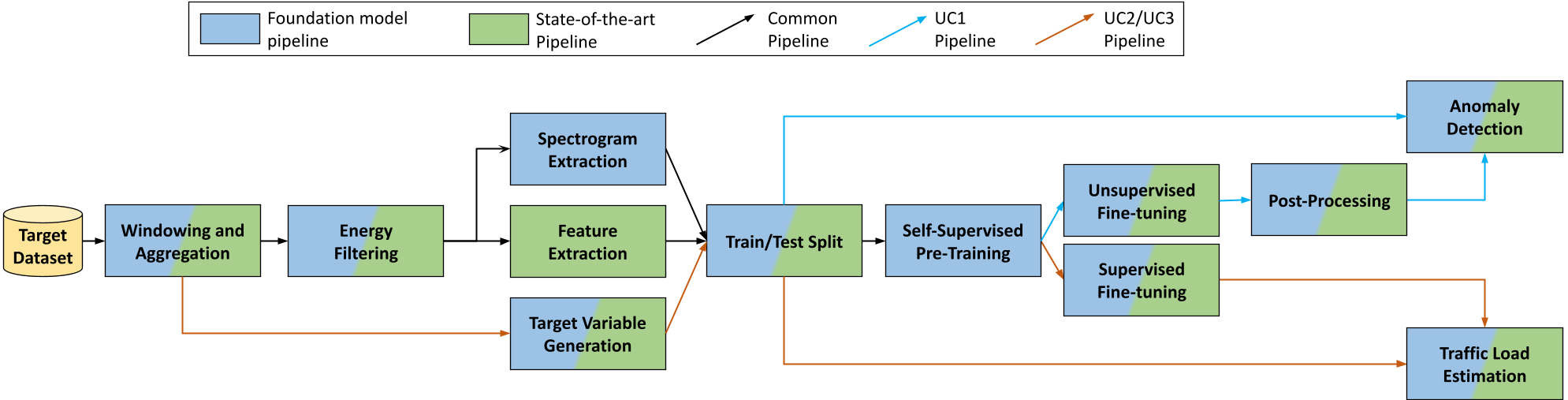}    
    \caption{Overview of the processing pipeline for traffic load estimation and anomaly detection. In blue are the blocks for our foundation model pipeline, and in green are the ones for SoA algorithms. Both colours are used for shared blocks. }
    \label{fig:pipeline}
\end{figure*}

This section presents our novel approach to vibration-based SHM leveraging self-supervised learning and Transformer foundation models. Firstly, we outline the data processing toolchain and train/test splitting (Section \ref{sec:pipeline}). Then, we delve into our model architecture (Section \ref{sec:model}) and training strategies and model exploration (Section \ref{sec:training}, Section \ref{sec:size_distillation}). 
The core objective of our work is to understand whether it is possible to pre-train a large Transformer model on substantial amounts of unlabeled data and then fine-tune it to realize different SHM tasks.
We consider both AD and TLE as downstream tasks to validate this hypothesis, focusing on all three Use Cases described in Sec.~\ref{sec:installation}. Specifically: 
\begin{itemize}
    \item AD is formulated as a ``one-class'' classification, discriminating between normal and anomalous structural states, similar to the approach of~\cite{9729869}. 
    \item  TLE is framed as a regression problem, predicting a scalar traffic value. For UC2, we split the estimation of light and heavy vehicle traffic into two separate tasks.
\end{itemize}

The complete pipeline is shown in Figure \ref{fig:pipeline}.

\subsection{Pre-processing Pipeline}
\label{sec:pipeline}

%
This section outlines the pre-processing applied before inputting data into our masked autoencoder. The pre-processing settings are consistent for training and validation data across all analyzed UCs, except when expressly indicated. 

The first pre-processing step is \textbf{windowing}, i.e., the division of the time-series data into time windows, which the algorithm can process. 
For UC1, we use 5s windows, as in the reference paper~\cite{9729869}, with a stride of 2s to increase the number of training samples.
For the other use cases, we enlarge the window to 60s since the reference study~\cite{burrello2022traffic} reported the best performance with this dimension, using a 2s stride in UC2, and 15s stride in UC3, given the extremely lower traffic on the viaduct (with a 2s stride, the information contained in successive windows was almost identical, while causing a linear increase in training time).
All windows are composed of uni-dimensional signals since we always employ either a single sensor or multiple sensors, but consider them separately. We also only consider the z-axis of the acceleration data.
Thus, each window is always a single vector of dimension $T = 100\times T_{win}$, where 100 is the acceleration sampling frequency in all three datasets. All time windows are normalized with mean and standard deviation.

After creating the windows, we apply \textbf{energy filtering}, i.e., we disregard as irrelevant those windows whose cumulative energy is under a certain threshold, $th =3.125\times(10^{-5})$ for UC1 and $th=1.25\times10^{-6}$ for UC2 and UC3. Energy thresholds are computed using the iterative process proposed in \cite{9729869}.

Lastly, building on the findings of \cite{huang2022masked}, which demonstrated that masked autoencoders achieve superior signal reconstruction in the frequency domain, we adopt a similar strategy in our approach. Specifically, we transform the raw acceleration data into its \textbf{spectrogram} representation, enabling the model to learn meaningful structural patterns through the reconstruction of masked spectrogram patches. Precisely, we feed the model with 100x100 time-frequency representations in the form of spectrograms. The first dimension corresponds to the temporal domain, while the second captures the frequency domain. While we notice that this window is big enough to obtain good task performance, we did not explore the downscaling of the spectrogram image to reduce the models' computational complexity, as the reduction of the input size has an impact only on the dimension of the last layer, whose size is negligible with respect to the whole model size.

For UC2 and UC3, we also need to generate the \textbf{target traffic variable}, while for UC1, samples are already separated between anomalous and normal based on time (see Sec.~\ref{subsec:data_uc1}). Despite being generated differently for the two use cases, respectively from an optical camera and a WiM system, the target variables for UC2 and UC3 are identical. Namely, as described in \cite{burrello2022traffic}, the target traffic load values for regression ($y$) in UC2 are set to:
\begin{equation}\label{eq:target}
    y = \frac{\sum_{t=1}^{T} (l_t = k)}{10}
\end{equation}
where $k=1$ for light vehicles and $k=2$ for heavy vehicles, and $\sum (l_t = k)$ indicates the number of samples that have label $k$ in that window. Notice that $y$ can also be a fractional number when the beginning/end of the window includes only a subset of a group of 10 consecutive samples sharing the same label.
For UC3, the only difference is the lack of separation between light vehicles and heavy vehicles. After pre-processing, data are split between \textbf{train and test} as described in Section \ref{sec:installation}, and summarized in Table \ref{tab:datasets}.

Lastly, we also apply a post-processing smoothing technique on the output of our trained model for UC1. Specifically, we label a window as anomalous or normal based on the \textit{median} over a variable size set of consecutive predictions. Namely, we experiment with median filters over $\{15, 30, 60, 120, 240\}$ consecutive windows.

\subsection{Foundation Model Architecture}
\label{sec:model}

\begin{figure*}
    \centering
    \includegraphics[width=0.99\textwidth]{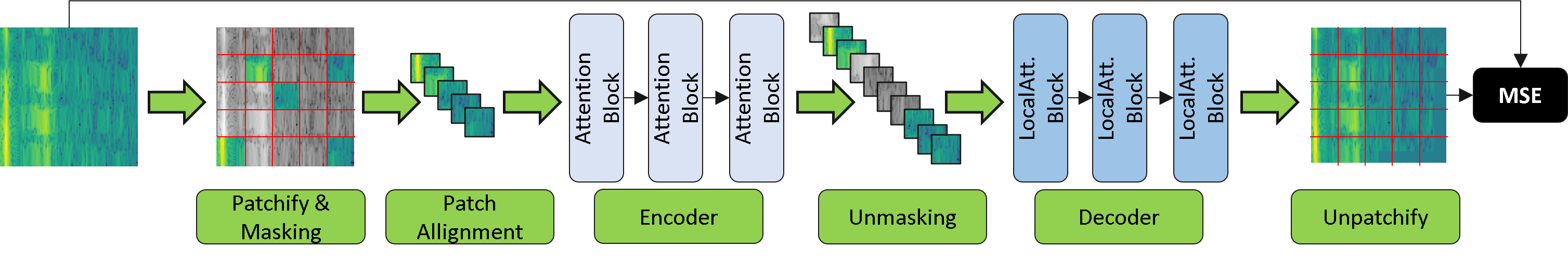}
    \caption{Architecture of our foundation model transformer-based masked autoencoder during pre-training.}
    \label{fig:base_model}
\end{figure*}
Our model architecture draws inspiration from the Masked Autoencoder proposed in \cite{huang2022masked}, which was originally introduced to learn self-supervised representations from audio spectrograms. The model uses a stack of standard Transformer blocks for both the encoder and the decoder, each consisting of multiple attention layers. As a backbone, we use ViT-Base (ViT-B). The model takes as input acceleration spectrograms divided into non-overlapped regular grid patches, with a configurable portion of them being masked during training. The complete architecture is shown in Fig.~\ref{fig:base_model}.

Architectural modifications are necessary to fine-tune our foundation model based on the downstream tasks considered. Namely, the model architecture remains unchanged when fine-tuning on UC1, because the task is framed as an outlier detection~\cite{9729869}: during training, the model learns the latent representation of normal data only, so that during testing, the decoder is expected to fail in accurately reconstructing data that deviates from its learned patterns. Accordingly, those samples whose reconstruction error (computed as detailed in Sec.~\ref{par:pretraining}) exceeds a user-defined threshold are classified as anomalies or outliers. On the other hand, the decoder is removed when fine-tuning on TLE tasks (UC2 and UC3), and a fully connected layer with a single neuron is appended directly to the latent representation produced by the encoder to produce the desired continuous traffic output defined in Eq.~\ref{eq:target}. The encoder is initialized with the weights learned during the self-supervised pre-training step, ensuring the transfer of the learned, general feature extraction. Additionally, to handle potential differences in input resolution, we interpolate the position embeddings as done in DeiT~\cite{deit}, preserving spatial consistency in learned representations.

\subsection{Training Phases}\label{sec:training}

The training of our masked autoencoder involves two main phases. Namely, the model is first pre-trained considering data from all three datasets and then fine-tuned on each single task. The two phases are detailed in the following, distinguishing the fine-tuning procedure for UC1 (AD) and for UC2/3 (TLE).
In both phases, we use an AdamW optimizer \cite{adamw} with different learning rates based on the training phase. For the pre-training phase, we use a learning rate of $0.25\cdot 10^{-3}$, while for the fine-tuning phase a learning rate of $0.25 \cdot 10^{-2}$ for UC1, $0.25\cdot 10^{-5}$ for UC2 and UC3, and a weight decay of $0.05$. 
Following what was presented in \cite{deit, beit}, we apply automatic gradient scaling to avoid exploding gradients.
During pre-training, we also progressively increase the learning rate from 0 to the pre-training learning rate value over the first 100 epochs to improve the convergence. After this, the learning rate is reduced following the half-cyle of the cosine function. We pre-train our foundation model for 200 epochs with a batch size of 128, then fine-tune it for 400, 500, and 200 epochs with batches of 64, 8, and 128 samples on UC1, UC2, and UC3, respectively.

\subsubsection{Pre-training}\label{par:pretraining}
The full encoder-decoder model, with the architecture of Fig.~\ref{fig:base_model} is initially pre-trained on a large amount of unlabeled data. We select pre-training data following the split described in Section \ref{sec:background}, combining the \textit{training sets of all three UCs}. Note that, although our data have labels, we do not use them in this phase, to simulate a scenario where the costly process of labelling all available samples is not affordable.

During pre-training, the non-overlapping spectrogram patches of each input sample are masked with probability $p$ (masking ratio) and aligned, before being passed to the transformer encoder. The latter then processes the non-masked patches to produce latent representations. Encoded patches are padded with trainable mask tokens and fed to the decoder, whose objective is to reconstruct the original spectrogram, including the missing parts. The encoder and decoder are jointly optimized to minimize the Mean Squared Error (MSE) between the reconstruction and the input spectrogram, averaged over masked patches. This pre-training is completely self-supervised and task-independent. Its objective is to learn robust latent representations of SHM data that allow the model to infer the underlying patterns in the data despite the missing information.

\subsubsection{Fine-tuning}\label{subsec:finetuning}
The details of the fine-tuning phase differ depending on the downstream task. For UC1 (AD), as mentioned above, we maintain the autoencoder structure of our model. Accordingly, the model is fine-tuned in an unsupervised way, using the same procedure just described, but considering only normal data from the UC1 dataset.
At test time, we use a threshold on the reconstruction error to identify anomalies. 
For both our models and the baseline PCA, we follow a procedure similar to~\cite{9729869} to determine it. Namely, the threshold is initialized to the average MSE on non-anomalous training data, and then progressively increased by a small step (1/1000th of the average), until 100\% specificity is achieved on a validation set, obtained as a 20\% sample of the training dataset.

For UC2, the pre-trained model is fine-tuned on acceleration data and video recording on the viaduct presented in Sec.~\ref{subsec:data_uc2}. Two separate fine-tunings are executed for light and heavy vehicle TLEs, starting from the same pre-trained model. For UC3, we consider the first half of the second day's data to fine-tune our foundation model, as described in \ref{subsec:data_uc3}. For these two downstream tasks, we follow a classic supervised fine-tuning procedure, where the model learns to predict the traffic load values rather than to reconstruct the input spectrograms. To fine-tune the encoder weights and train the added fully-connected layer, we use an initial lower learning rate of $0.25\cdot 10^{-5}$ that is updated with a half-cycle cosine decay schedule. As a loss function, we use the MSE between the newly inserted fully connected layer's output prediction and the ground truth target variable. 

\subsection{Model Exploration and Knowledge Distillation}
\label{sec:size_distillation}
To select our model's hyperparameters, we do a search over the encoder/decoder depth and heads, the embedding dimensions, and the masking probability $p$ of the masked autoencoder. To make this search manageable, we sample 20\% of the pre-training data, further splitting them into 80\% for training and 20\% for validation. The task is the same as the pre-training, i.e., reconstructing the masked input data. The best configuration is selected based on the lowest error achieved on the validation set.
We employ Bayesian optimization with the open-source Optuna framework\footnote{https://optuna.org}, fixing $p=0$. Then, we explore $p$ through a grid search with a step of 0.2. Table 1 summarizes the search space, with the optimal configuration highlighted in bold.

\begin{table}[t]
\centering
\caption{Explored space of architectural hyperparameters for the Transformer encoder-decoder}
\resizebox{0.4\columnwidth}{!}{%
\begin{tabular}{c|c}
\textbf{Parameters} & \textbf{Values explored} \\ \hline
Encoder depth & $[1, 3, 5, \textbf{7}]$ \\
\# Encoder Heads & $[\textbf{8}, 12, 16]$ \\
Encoder embedding & $[1536, \textbf{768}, 384, 192]$ \\
Decoder depth & $[\textbf{2}, 4, 6, 8]$ \\
\# Decoder Heads & $[8, \textbf{16}, 24]$ \\
Decoder embedding & $[1024, \textbf{512}, 256, 128]$ \\ \hline
Masking ratio & $[0,0.2, 0.4, 0.6, \textbf{0.8}]$ \\
\end{tabular}%
}
\label{tab:params}
\end{table}
After this step, we also define a family of smaller models to explore the possibility of reducing the computational burden with minimal accuracy loss, targeting the deployment on sensor nodes with an extremely low memory budget. Starting from the frozen configuration of Table \ref{tab:params}, we systematically reduce the embedding dimensions of both the encoder and decoder, starting from the baseline configuration $(e_{dim},d_{dim})=(768,512)$ and scaling them down by a factor of $2$ until reaching $(e_{dim},d_{dim})=(24,16)$.
These new models are pre-trained on all datasets and subsequently fine-tuned on the three downstream tasks, following the same procedure described in the previous section. Additionally, to inherit our big model performance, we considered a modified fine-tuning procedure enhanced with Knowledge Distillation (KD). In this setup, considering the baseline model as a teacher, the student utilizes a multi-objective fine-tuning loss function composed of two contributions: the first one, $\mathcal{L}_{task}$, is the MAE computed on student prediction against the ground truth, while the second one, $\mathcal{L}_{KD}$, is the Root Mean Squared Error (RMSE) between student and teacher predictions. Through an empirical approach, we assessed that evenly balancing these two contributions leads to faster convergence of the overall loss, whose function becomes:
\begin{equation}
    \mathcal{L} = 0.5 \cdot \mathcal{L}_{task}\left(y_{s}, y_{true}\right) + 0.5 \cdot \mathcal{L}_{KD}\left(y_{s}, y_{t}\right)
\end{equation}
where $y_s$, $y_t$, and $y_{true}$ are the student, the teacher, and the true predictions, respectively. 
Even when using KD in fine-tuning, for all UCs, student models are always first pre-trained following the self-supervised procedure of Sec. \ref{par:pretraining}. 
\section{Results}\label{sec:results}
\begin{table}[ht]
\caption{Information about the use cases' dataset. Abbreviations: AD: anomaly detection, TLE: traffic load estimation.}
\label{tab:datasets}
\resizebox{\columnwidth}{!}{\begin{tabular}{l|lllllll}
    & Task              & N. of Samples {[}Train / Test{]} & Time {[}Train / Test{]} & Data Dimension & Data Stride & N.of Sens. & Label                      \\ \hline
UC1 & AD &      302.4k / (172.8k N., 172.8k A.)                            &  7d / (4d N., 4d A.)                       &      5s $\times$ 100Hz           &      2s $\times$ 100Hz       & 1            & Normal / Anomaly           \\
UC2 & TLE               &      651 / 279                            &     21:42 / 9:18 min:sec                    &     60s $\times$ 100Hz           &   2s $\times$ 100Hz    & 1            & \# of Light/Heavy Vehicles \\
UC3 & TLE               &               699.9k / 50k                   &   36 hours / 6 hours                      &    60s $\times$ 100Hz           &   15s $\times$ 100Hz & 142          & \# of Vehicles             \\ \hline
\end{tabular}
}
\end{table}

In this section, we first introduce the datasets used for training, validation and testing, summarized in Table~\ref{tab:datasets}. Notice that, identically to state-of-the-art works, data splitting is tailored for each task, considering the fundamental differences between anomaly detection (UC1) and TLE (UC2 and UC3).
Then, we introduce the comparison baselines and detail the results of our foundation models on the three use cases. We also quantify the impact of self-supervised pre-training on the performance of our models. Lastly, we analyze the trade-off between model size and performance and compare regular and KD-enhanced fine-tuning. In all experiments except the last one, we consider our largest masked autoencoder version, i.e., $(e_{dim}, d_{dim})=(768, 512)$.

\subsection{Datasets}
\paragraph{UC1}\label{subsec:data_uc1}
the employed dataset is the same as the one described in \cite{9729869}, comprising a total of 15 days of continuous monitoring of the viaduct.
For our analysis, we consider only the central sensor of the chain, which is the one most influenced by the viaduct vibration. 
The data include measurements from 4 days before the maintenance intervention, labeled as anomalies, and two intervals of 4 and 7 days respectively, after the intervention, labeled as normal data.
We select as the test set the two intervals of 4 days to have a perfect balance between anomalies and normal data.
The remaining 7 days of normal data are used to train the AD algorithm (and the AD threshold).
Note that anomalies are not used during training given that, for this task, we only consider unsupervised models trained to identify anomalies as outliers. 
\paragraph{UC2}\label{subsec:data_uc2}
we analyze only section 10 of the viaduct, where an optical camera provides ground truth labels for TLE. This section includes seven sensors mounted on different tendons, but we use only the central one, which experiences the highest vibrations. The dataset consists of 31 minutes of recorded z-axis acceleration data and a synchronized video, split 70:30 for training (21 minutes and 42 seconds) and testing (9 minutes and 18 seconds). Data were collected between 8:00 a.m. and 9:00 a.m., with the camera operating at 10 FPS. Labels were assigned based on vehicle crossings, using the video: a value of 1 for light vehicles, 2 for heavy vehicles, and 0 otherwise. Given the accelerometer’s 100 Hz sampling rate and the camera’s 10 FPS, each label was assigned to a set of 10 consecutive acceleration samples. The final TLE label was computed as the sum of 1s and 2s in the window, divided by 10, following \cite{burrello2022traffic} (see Sec.~\ref{sec:pipeline}).

\paragraph{UC3}\label{subsec:data_uc3}
we utilize two days of data to train and evaluate our algorithms.
%
We use the first complete day's data, spanning from 00.00 to 23.59, and covering a variety of traffic conditions only as pre-training data for the foundation model. We then use the first half of the following day (00.00 to 12.00) as additional training data for supervised fine-tuning on this specific task. Finally, the last half day's data is reserved for testing.
It's important to note that, in the 600 m between the WiM and the bridge, vehicle speeds may vary, with potential lane changes, vehicle overtaking, and overlapping. As a result, vibrations on the bridge may not be entirely correlated with the data collected from the WiM, which can unavoidably lead to a loss of accuracy in this analysis.

\subsection{Comparison Baselines}
We use different baseline algorithms depending on the UC, given the difference in the solved task, i.e., AD vs. TLE. 
For UC1, we compare against the top-performing model of \cite{9729869}, a PCA that considers a window dimension of 5 seconds (500 samples at 100 Hz) and a post-processing median filter with varying length to aggregate consecutive predictions. We underline that this solution has already been shown to outperform convolutional and fully connected deep autoencoders, as demonstrated in~\cite{9729869}.

For UC2 and UC3, we compare with the five supervised algorithms proposed in \cite{burrello2022traffic}, i.e., the work that first introduced the UC2 dataset.
It is worth noticing that this is the task with the largest dataset available for training: this could further explain the superiority of foundation-model-based approaches, which are expected to thrive when more data are available, and, at the same time, justify the limits of SoA approaches, as their shallow nature may be unsuited to handle this complexity.
Namely, we consider: i) a K-Nearest Neighbors (k-NN); ii) a Linear Regressor (LR); iii) a Random Forest (RF) trained with the variance reduction criterion~\cite{cart}; iv) a Support Vector Regressor (SVR); v) a Multi-Layer Perceptron (MLP) with ReLU activations, trained with the \textit{Adam} optimizer, a learning rate of $10^{-3}$ and a mini-batch size of 200. In addition to that, we consider two alternative deep learning architectures, a Temporal Convolutional Network (TCN) and a Long Short-Term Memory (LSTM) network. We implement the TCN architecture of \cite{tcn} composed of four 1D convolutional layers with increasing dilation factor from first to last block, followed by a regression head consisting of two linear layers with ReLU activation and dropout. For the LSTM, we consider a variation of the architecture from \cite{lstm} reducing its layers from 4 to 3 due to overfitting, and changing the hidden size to match the one of our Transformer Encoder (768). Also in this case, the regression head is composed of two linear layers with ReLU activation and dropout. On these last two architectures, we test both direct supervised training on the task-specific dataset, as described in Section \ref{sec:training}, and a two-phase foundation model approach including self-supervised pre-training on all the un-labeled data followed by supervised fine-tuning. For the latter, we use the same masked autoencoder framework of Figure~\ref{fig:base_model}, except that the Transformer-based Encoder and Decoder are replaced by TCN and LSTM respectively.
We build the Decoders architectures symmetrically to the Encoders: for the TCN, we replace convolutions with up-sampling convolutions, while for the LSTM, we add a linear layer to project the output of the encoder to the decoder embedding dimension (512), once again matched to the one of our Transformer.
Section~\ref{res:uc2_uc3} only reports the results with pre-training, which has proven superior for both types of model.
A summary of the baseline models and their hyperparameters is shown in Table \ref{tab:models}.
\begin{table}[t]
\centering
\caption{Models employed in our experiments.}
\label{tab:models}
\begin{tabular}{lll}
\multicolumn{1}{l|}{Model}                        & \multicolumn{1}{l|}{Task}              & Hyperparameters             \\ \hline
\multicolumn{3}{c}{State-of-the-art}                                                                                     \\ \hline
\multicolumn{1}{l|}{LR}          & \multicolumn{1}{l|}{TLE}               & -                           \\
\multicolumn{1}{l|}{RF}                & \multicolumn{1}{l|}{TLE}               & Depth = 200, Trees = 30 \\
\multicolumn{1}{l|}{k-NN}        & \multicolumn{1}{l|}{TLE}               & k = 7                       \\
\multicolumn{1}{l|}{MLP}       & \multicolumn{1}{l|}{TLE}               & Layers = 3, Neurons = 100   \\
\multicolumn{1}{l|}{SVR}     & \multicolumn{1}{l|}{TLE}               & Kernel = RBF, C = 10.0, $\epsilon$=0.1      \\
\multicolumn{1}{l|}{PCA} & \multicolumn{1}{l|}{AD} & CF = 32                     \\ \hline
\multicolumn{3}{c}{Our Work}                                                               \\ \hline
\multicolumn{1}{l|}{Masked autoencoder} & \multicolumn{1}{l|}{TLE, AD} & See Section \ref{sec:methodology}                     \\ \hline
\end{tabular}
\end{table}

\subsection{Anomaly Detection (UC1) Results}

\begin{figure}
    \centering
     \includegraphics[width=0.7\columnwidth]{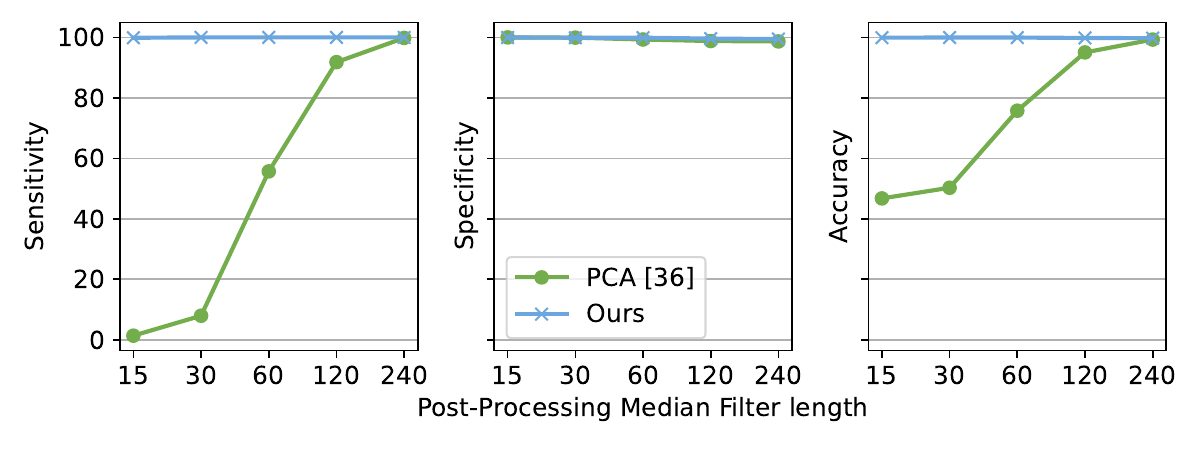}
    \caption{SoA PCA \cite{9729869} vs. masked autoencoder performance while varying the smoothing post-processing median filter length.}
    \label{fig:anomaly_bars}
\end{figure}

\begin{figure}
    \centering
    \includegraphics[width=0.7\textwidth]{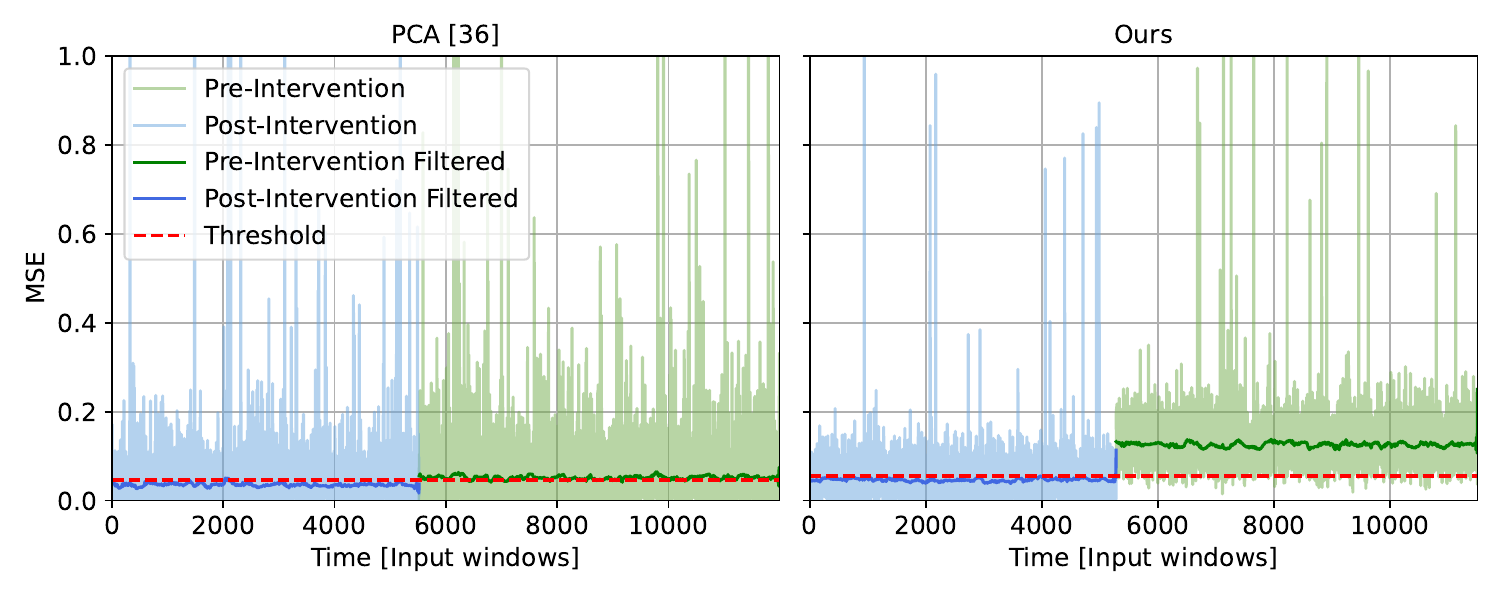}
    \caption{Reconstruction error of the PCA~\cite{9729869} and our proposed model, for normal and anomalous windows}\label{fig:anomaly_MSE}
\end{figure}

We evaluate AD models in terms of \textit{specificity}, computed as the proportion of correctly classified normal windows, \textit{sensitivity}, i.e., the proportion of correctly classified anomalous windows, and \textit{accuracy}, i.e., the overall ratio of correctly classified windows, providing a general measure of performance.
Figure \ref{fig:anomaly_bars} shows the comparative performance of PCA and our foundation model while varying the number of windows outputs fed to the median filter as post-processing. The graph highlights the increasing trend of accuracy, sensitivity, and specificity as the dimension increases for both methods. However, we can observe a significant improvement in all evaluation metrics of our model compared to the PCA, especially for short time windows. For example, when applying the median smoothing to 60 windows, our masked autoencoder achieves an accuracy of 99.92\%, while the PCA only reaches 75.76\%. To match the accuracy of our model, the PCA needs a post-processing median filter of 240 windows, i.e., 4x larger. The sensitivity shows a similar trend: for a 60-windows smoothing, our model reaches 99.9\%, compared to 55.68\% of the PCA.

Having a shorter post-processing median filter translates into a significantly lower anomaly detection latency. For example, given that each window is 5s long, a median filter over 240 windows, as required by the PCA, corresponds to a latency of 20 minutes before an anomaly can be signalled. In contrast, our method reduces this latency to 5 minutes.

Additionally, the specificity values demonstrate the masked autoencoder's superiority in correctly identifying normal data instances. We achieve an almost perfect specificity (min. 99.97\%) for all time window dimensions, whereas PCA shows lower specificity values, ranging from 96.21\% to 99.87\%.

To visualize the performance of our model compared to the PCA, Figure \ref{fig:anomaly_MSE} shows the reconstruction error (MSE) of normal and anomalous windows. The darker line shows the output of median-smoothing over 120 windows. 
From the figure, we can notice that the PCA reconstruction error is more noisy, and, after the smoothing post-processing, it shows a very small margin between normal and anomalous signals. On the other hand, our approach shows a much more pronounced difference between pre and post-intervention data, thus being able to more robustly tell apart normal and anomalous windows.

\subsection{Traffic Load Estimation (UC2 and UC3) Results}\label{res:uc2_uc3}
%
To compare the performance of TLE models, we consider five different metrics: \textit{Mean Squared Error (MSE)}, which measures the average squared difference between predicted and true values; \textit{Mean Absolute Error (MAE)}, which computes the average absolute difference between predictions and ground truth; \textit{R$^2$ score}, which quantifies how well the model explains the variance in the data, with values closer to 1 indicating better performance; \textit{Mean Squared Percentage Error (MSE\%)}, and \textit{Mean Absolute Percentage Error (MAE\%)}, obtained respectively by dividing MSE and MAE by the average predicted value on the test set, allowing for a relative error assessment with respect to the true values.

\begin{table*}[t]
\centering
\footnotesize
\caption{Comparison between the algorithms of \cite{burrello2022traffic} and the masked autoencoder proposed in our work on UC2. The comparison uses the features extraction of~\cite{burrello2022traffic} for the baselines and raw data for our method. Bold = best result, underline = second best.}
\label{tab:algorithm_comparison}
\resizebox{\columnwidth}{!}{\begin{tabular}{lllllllllll}
\multicolumn{1}{l|}{}                   & \multicolumn{5}{c|}{Heavy Vehicles}                      & \multicolumn{5}{c}{Light Vehicles}   \\
\multicolumn{1}{l|}{Model}              & MSE  & MAE  & R$^2$ & MSE\% & \multicolumn{1}{l|}{MAE\%} & MSE  & MAE  & R$^2$ & MSE\%  & MAE\% \\ \hline
\multicolumn{11}{c}{State-of-the-art}                                                                                                     \\ \hline
\multicolumn{1}{l|}{SVR}                 & \underline{0.35} & 0.42 & \underline{0.91}  & \underline{10.76} & \multicolumn{1}{l|}{12.85} & 2.23 & 1.05 & 0.76  & 37.03  & 17.50 \\
\multicolumn{1}{l|}{RF}                 & 0.38 & \underline{0.32} & 0.91  & 11.76 & \multicolumn{1}{l|}{\underline{9.66}}  & \underline{1.50} & \textbf{0.73} & \underline{0.84}  & \underline{25.02}  & \textbf{12.10} \\
\multicolumn{1}{l|}{MLP}                & 0.53 & 0.53 & 0.87  & 16.37 & \multicolumn{1}{l|}{16.37} & 2.99 & 1.35 & 0.67  & 49.69  & 22.44 \\
\multicolumn{1}{l|}{kNN}                & 0.37 & 0.41 & 0.91  & 11.47 & \multicolumn{1}{l|}{12.45} & 2.26 & 1.09 & 0.75  & 37.59  & 18.14 \\
\multicolumn{1}{l|}{LR}                & 1.19 & 0.84 & 0.71  & 36.57 & \multicolumn{1}{l|}{25.82} & 7.34 & 2.14 & 0.20  & 121.94 & 35.67 \\ \hline
\multicolumn{11}{c}{Our Work}                                                                                                             \\ \hline
\multicolumn{1}{l|}{Masked Autoencoder} & \textbf{0.11} & \textbf{0.25} & \textbf{0.97}  & \textbf{3.42}  & \multicolumn{1}{l|}{\textbf{7.88}} & \textbf{0.95} & \underline{0.75} & \textbf{0.90}  & \textbf{15.92}  & \underline{12.44} \\ \hline
\end{tabular}
}
\end{table*}

\subsubsection{UC2}
Table \ref{tab:algorithm_comparison} reports the results in terms of these metrics obtained on UC2 for light and heavy vehicles. Considering the latter, we outperform the best baseline models from~\cite{burrello2022traffic} (SVR and RF) in all metrics. However, our foundation model reduces the MSE by 3.2x, the MAE by 1.3x, and improves the R$^2$ of +0.06. 
We impute this to the increased capacity of our foundation model in combination with the transformer architecture, which permits to better leverage the non-linear and higher-order interactions within acceleration data that correlate with traffic flow, while current SoA machine learning methods exhibit limitations in capturing these intricate relationships.
Noteworthy, the other deep learning architectures tested do not attain satisfactory performance, being inferior to classical machine learning algorithms
Additionally, notice how there is no single best SoA model, while our approach convincingly overcomes them all in every considered metric.
For the prediction of the light vehicle traffic, our foundation model outperforms the SoA in terms of MSE (1.6x reduction) and R$^2$ (+0.06) and is the second best in terms of MAE (1.03x increase over the best baseline model, the RF).
Importantly, having a significantly lower MSE and a slightly higher MAE means that our approach does not output predictions that are far off the ground truth, as the quadratic function of the MSE penalizes major mistakes. 
Thus, it could be more amenable for a real-world deployment than the baseline RF.

\begin{figure}
    \centering
    \includegraphics[width=0.7\textwidth]{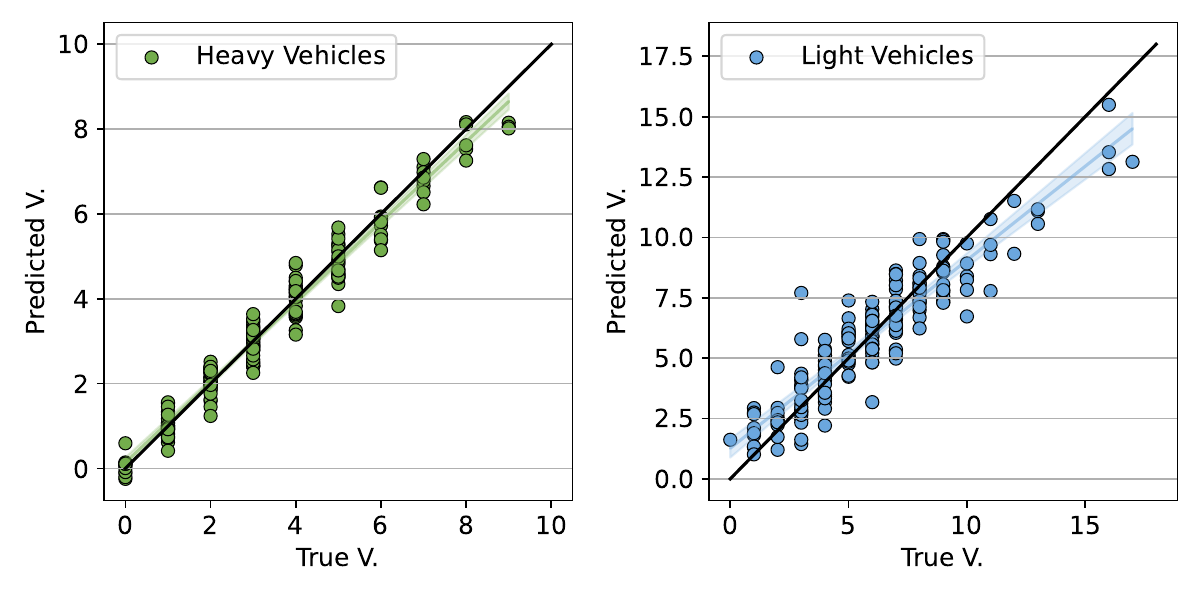}
    \caption{True vs predicted values of the TLE regression task for heavy and light vehicles. The black line represents the ideal prediction (true value = predicted value). The coloured straight lines are the linear interpolation of the predicted samples, and the light-coloured area corresponds to the 99.5\% confidence interval of the interpolation.}
    \label{fig:roccaprebalza_regression}
\end{figure}

Figure~\ref{fig:roccaprebalza_regression} plots the ground truth target variable against our model's predictions to further highlight our approach's effectiveness. As shown, most predictions are close to the target without noticeable outliers. The black line represents the ideal prediction, where true values exactly correspond to the predicted values. 
The coloured straight lines correspond to the linear interpolation of the predicted samples, while the light-coloured area corresponds to the 99.5\% confidence interval of the interpolation. 
For the heavy vehicle case, the confidence interval almost overlaps with the ideal prediction across traffic conditions (i.e., x-axis values), further validating the effectiveness of our approach in this scenario.
For light vehicles, when facing high traffic, the model's confidence interval slightly deviates from the ideal prediction, underestimating the traffic on the viaduct. This may be a consequence of the lower weight of light vehicles. The latter generate vibrations with lower amplitudes on the viaduct, where vibrations caused by multiple vehicles can be easily mistaken for a single vehicle in light traffic. Further, the model is biased towards lower-value TLE estimations, given the scarcity of high-traffic window samples.

\subsubsection{UC3}
\begin{table}[t]
\caption{Comparison between the algorithms of \cite{burrello2022traffic} and the masked autoencoder proposed in our work on the UC3. The comparison uses all the extracted features for the baselines and raw data for the masked autoencoder.}
\label{tab:sacertis_results}
\centering
\begin{tabular}{llllll}
\multicolumn{1}{l|}{Model}              & MSE   & MAE  & R$^2$ & MSE\%   & MAE\%  \\ \hline
\multicolumn{6}{c}{State-of-the-art}                                              \\ \hline
\multicolumn{1}{l|}{LR}                 & 20.53 & 2.05 & -14.40 & 1091.02 & 109.03 \\
\multicolumn{1}{l|}{RF}                 & 2.23  & 1.10 & -0.67  & 118.48  & 58.30  \\
\multicolumn{1}{l|}{kNN}                & 1.34  & 0.90 & -0.01  & 71.14   & 47.81  \\
\multicolumn{1}{l|}{MLP}                & \underline{1.20}  & \underline{0.85} & \underline{0.10}  & \underline{63.89}   & \underline{45.39}  \\
\multicolumn{1}{l|}{SVR}                & 1.39  & 0.91 & -0.04  & 73.77   & 48.19  \\ \hline
\multicolumn{6}{c}{Our Work}                                                      \\ \hline
\multicolumn{1}{l|}{Masked Autoencoder} &     \textbf{0.62 } &  \textbf{0.57}    &  \textbf{0.54}    &    \textbf{32.93 }    &     \textbf{30.13}   \\ \hline
\end{tabular}
\end{table}
 
The results of UC3 are shown in Table \ref{tab:sacertis_results}. We compare our approach with the same models considered for UC2. However, since there is no previous literature on this dataset, we re-train all baseline models with the same data split used for our foundation model while keeping the same hyperparameter settings of~\cite{burrello2022traffic}. Results show that our foundation model overcomes state-of-the-art performance on all metrics, improving the MSE and MAE of the second-best, the LSTM, by 0.03 and 0.02, respectively, while increasing the R$^2$ score by 0.03 as well. We observe two general trends: first, the performance of UC3 is worse compared to UC2's on every metric and for every algorithm. Second, the second-best performing model is the LSTM, trained with the same foundation-model approach as our transformer. This is in strong contrast with UC2, where both the other deep learning based models, TCN and LSTM, reached low performance.

The observed performance discrepancy for both state-of-the-art and our proposed approaches on UC3 compared to UC2 (and their performance ranking) can be attributed to two key factors. 
First, as anticipated in Section \ref{sec:installation}, the label assignment procedure likely introduces noise into the training data. The physical separation of the WiM system from the bridge affects the correlation between bridge vibrations and WiM collected data. This misalignment between predicted and actual bridge conditions may partially explain the performance drop observed in this use case.
Sensor placement may also affect task difficulty. In fact, unlike UC2, sensors are positioned on two external beams per span, which might lead to reduced sensitivity to vehicle-induced vibrations. 
Secondly, UC3 comprises roughly 750,000 samples, over three orders of magnitude more than UC2’s dataset, 1,000 samples. DNN models benefit more than classic ML ones from large-scale data, reaching, as in the case of the LSTM, significantly better performance compared to UC2.

\subsection{Impact of pre-training}
\begin{figure}
    \centering
    \includegraphics[width=0.6\textwidth]{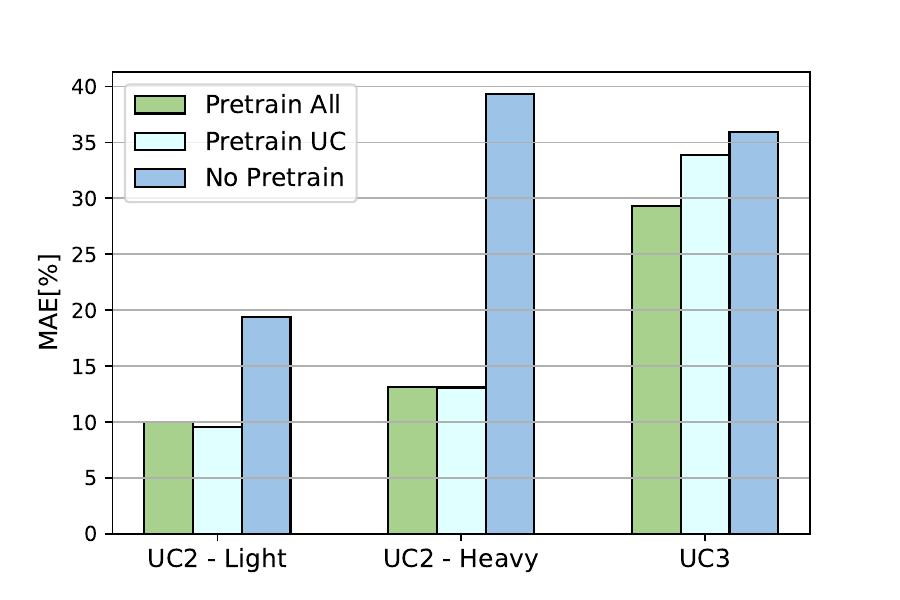}
    \caption{Impact of different pre-trainings on masked autoencoder outcome. }
    \label{fig:pretraining}
\end{figure}

In this section, we demonstrate that self-supervised pre-training is a fundamental contributor to the performance of our foundation models. Considering UC2 and 3, we apply three different training setups: 
\begin{itemize}
    \item \textbf{No Pretrain}: the pre-training step is completely disregarded, and the model is directly trained in a supervised way considering only UC-specific data;
    \item \textbf{Pretrain UC}: the model is first pre-trained and then fine-tuned only considering data of the specific use case. Thus, the \textit{same training data} are applied during the masked autoencoder self-supervised pre-training training and during supervised fine-tuning (after replacing the decoder with the classification head);
    \item \textbf{Pretrain All}: the model is pre-trained on all three datasets (combining the respective training sets) and fine-tuned on UC-specific data.
\end{itemize}
The test MAE\% obtained by our foundation model in the three cases are reported in Figure~\ref{fig:pretraining}. As shown, \emph{No Pretrain} always performs worse than the other two setups, underlying the usefulness of self-supervised pre-training. 
Quantitatively, \emph{Pretrain All} reduces the MAE\% by 1.94$\times$, 3$\times$ and 1.2$\times$ on UC2-Light, UC2-Heavy and UC3, respectively, compared to \emph{No Pretrain}. As expected, the benefits of pre-training are more evident for UC2, which has the smallest dataset.
Moreover, pre-training the foundation model on all three datasets shows the best results for UC3 (with \textit{Pretrain All} achieving 1.2$\times$ lower MAE\% compared to \textit{Pretrain UC}), while no clear advantage is evident for UC2. 
This is in accordance with the observation of the previous section that UC3 is 
the most difficult downstream tasks among the three,
thus benefitting more from high-quality and generalizable representations.
These findings indicate that the effectiveness of this approach might be further amplified by leveraging substantially larger and more diverse datasets for pre-training.

\subsection{Model size optimization and Knowledge Distillation}

\begin{figure}
    \centering
    \includegraphics[width=0.9\textwidth]{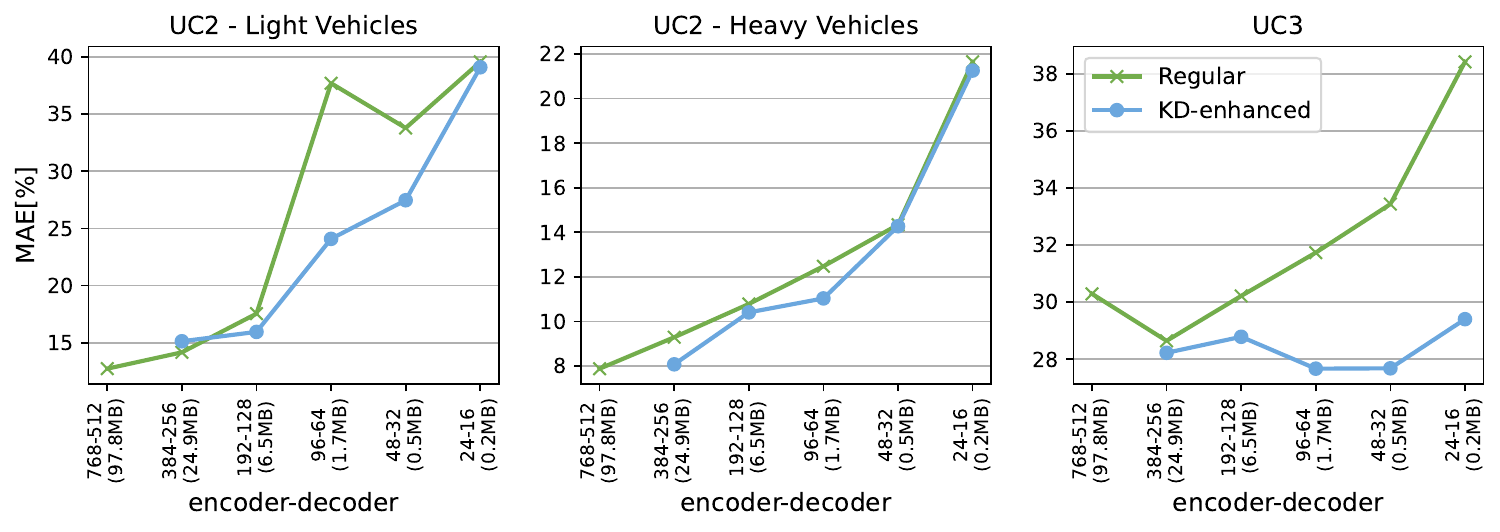}
    \caption{MAE\% comparison between regular and KD-enhanced fine-tuning. All graphs share the same x-axis.}
    \label{fig:regular_vs_fine-tuning}
\end{figure}

We also analyze the trade-off between model size and performance.
For all UCs, following the findings of the previous section, we use a \textit{Pretrain All} scheme, and we compare how regular supervised fine-tuning stands against KD-enhanced fine-tuning.
We focus on TLE tasks because, for UC1, we were able to obtain the same performance shown in Figure \ref{fig:anomaly_bars} for the largest transformer, with a model of just $0.2$MB, with $(e_{dim}, d_{dim}) = (24, 16)$. Results for UC2 and UC3 are shown in Figure \ref{fig:regular_vs_fine-tuning}. For clarity's sake, we present results only in terms of MAE\%, as other metrics follow similar trends. 
As shown, KD helps improve the performance of the models in almost all configurations, with the exception of a few cases related to UC2, especially in the low-size regime, where it provides marginal advantages. This is motivated by the fact that student networks that are too small cannot imitate the predictions of much larger teachers. Hence, the KD loss only adds noise to the training signal. At most, on UC2, KD improves the MAE by 1.6$\times$ on light vehicles, at $(e_{dim}, d_{dim})=(96, 64)$, and by 1.15$\times$ on heavy vehicles, at $(e_{dim}, d_{dim})=(384, 256)$, compared to regular fine-tuning. In the latter case, the KD-enhanced model reaches a MAE\% that is just 1.2$\times$ higher than the teacher while being 3.9$\times$ smaller.

The most interesting results are obtained on UC3. Here, we first note that reducing the model size to $(e_{dim}, d_{dim})=$(384-256) is beneficial for performance (5\% reduction for a 3.9$\times$ smaller size) even without KD.
Moreover, we can achieve very competitive results even with smaller models thanks to our  KD-enhanced fine-tuning setup, which is very effective. The most interesting result is achieved at $(e_{dim}, d_{dim})=(48,32)$, which further reduces the MAE\% by 8.6\% compared to the baseline, with a model whose parameters only require $0.5$MB, i.e., $195\times$ smaller than the original one.

These results show the effectiveness of KD in reducing model size while preserving as much as possible the performance of foundation models from the perspective of real-world deployment resource-constrained devices. The two smallest student configurations require less than 0.7 MBs and could be even deployed on the data collection nodes mentioned in Section~\ref{sec:installation}, which are equipped with 1 MB of Flash memory. However, it must be underlined that, on UC2, these downsized models remain hindered by performance gaps that might be too large depending on application-specific requirements.

\subsection{Deployment results}
\begin{table*}[t]
\centering
\caption{Deployment results of for a single inference step on NVIDIA Jetson Nano. UC2/UC3 results consider a 60 s input window.}
\label{tab:latency}
\resizebox{0.9\textwidth}{!}{%
\begin{tabular}{c|cc|ccccccccc}
\textbf{Task} & \multicolumn{2}{c|}{UC1} & \multicolumn{9}{c}{UC2/UC3} \\ \hline
\textbf{Model} & PCA & Ours & LR & RF & k-NN & MLP & SVR & TCN & LSTM & Ours & \begin{tabular}[c]{@{}c@{}}Ours (distilled)\end{tabular} \\ \hline
\textbf{Params} & $<$ 100k & 36 M & $<$ 100k & $<$ 100k & $<$ 100k & $<$ 100k & $<$ 100k & 11 M & 42 M & 25 M & 130k \\
\textbf{Latency {[}ms{]}} & 14.81 & 169.36 & 0.26 & 0.72 & 4.63 & 0.66 & 0.52 & 127.91 & 1533.26 & 164.93 & 40.12
\end{tabular}
}
\end{table*}
While further research is necessary to port these models on SHM sensor nodes, in this section we show that running them \textit{in real-time} on an edge gateway (present in virtually every sensor node installation, including the ones we considered) is feasible without any loss in accuracy.
Table \ref{tab:latency} summarizes the deployment results of our architectures on an NVIDIA Jetson Nano, as a proxy for a typical AI-oriented SHM gateway. The module features an NVIDIA Maxwell architecture with 128 CUDA cores and a Quad-core ARM Cortex-A57 MPCore processor, equipped with 4 GB of 64-bit LPDDR4 memory and 16 GB of storage. Inference times were measured across all architectures to assess their real-time feasibility in deployment scenarios, averaging results across 1000 test samples using PyTorch 1.1.0, with model and data in full precision (fp32); because the deployed models are identical to those evaluated offline, their accuracy remains unchanged.

Among all the use cases, the biggest transformer model is the one for UC1 (which includes the Decoder), with 36M parameters: in fp32, this model occupies a total of 144 MB, easily fitting the target gateway's memory.
This model achieves an inference latency of 169.36 ms. While the PCA-based anomaly detector is faster, our transformer is still well below the 2-second constraint imposed by the stride between successive input windows.

For UC2 and UC3, both using the same model where the Decoder has been replaced by a fully connected layer, we achieve 164.93 ms latency, well below the real time constraints of 2 / 15 seconds of UC2 / UC3. As expected, the distilled model is significantly faster, requiring only 40 ms. Classical ML baselines exhibit latencies below 1 ms, which are however unnecessary in this context.
\section{Conclusions}\label{sec:conclusions}
We proposed a task-independent foundation model for SHM applications, which can achieve state-of-the-art performance on two tasks, Anomaly Detection and Traffic Load Estimation, considering three different datasets. Our transformer-based masked autoencoder learns powerful representations of vibration data gathered by accelerometers during the pre-training step and can be then effectively fine-tuned on the specific downstream task.
Even if limited to a few SHM tasks, we believe that our work points to a new and exciting direction in ML for SHM research.
Future works will include using larger and more diverse pre-training datasets, as our results already suggested this could lead to possible benefits, investigating more SHM tasks involving different civil infrastructures, as well as experimenting more thoroughly with model hyper-parameters.

\section{Acknowledgment}
This publication is part of the project PNRR-NGEU which has received funding from the MUR – DM 117/2023.
We thank Sacertis Ingegneria Srl for providing the data for this research.
\bibliographystyle{IEEEtran}
\bibliography{lib_new}

\end{document}